\documentclass{llncs}
\usepackage[T1]{fontenc}
\usepackage[utf8]{inputenc}
\usepackage{lmodern}
\usepackage{amsmath}
\usepackage{algorithm}
\usepackage{amssymb}
\usepackage{algorithmic}
\usepackage{graphicx}
\usepackage{tikz}
\usetikzlibrary{arrows,shapes,shadows,positioning,fadings,decorations,decorations.pathmorphing}
\tikzstyle{block}=[draw opacity=0.7,line width=1.4cm]
\newcommand{\node}{v}

\title{Graph Kernels exploiting Weisfeiler-Lehman Graph Isomorphism Test Extensions}
\author{Giovanni Da San Martino \and Nicol\`o Navarin \and Alessandro Sperduti} 
\institute{Department of Mathematics, University of Padova \\\email{\{dasan,nnavarin,sperduti\}@math.unipd.it}}
\begin{document}

\maketitle

\begin{abstract}
In this paper we present a novel graph kernel framework inspired the by the Weisfeiler-Lehman (WL) isomorphism tests. 
Any WL test comprises a relabelling phase of the nodes based on test-specific information extracted from the graph, for example the set of neighbours of a node. 
%
We defined a novel relabelling and derived two 
kernels of the framework from it. The novel kernels are very fast to compute and achieve state-of-the-art results on five real-world datasets.
\end{abstract}
\section{Introduction}
In many real world learning problems, input data are naturally represented as graphs~\cite{springerlink:10.1007/s10115-007-0103-5}\cite{Weislow1989}.
A typical approach for solving machine learning tasks on structured data is to project the input data onto a vectorial feature space and then perform learning on such space. Ideally, a good projection should ensure non-isomorphic data to be represented by different vectors in feature space, i.e. to be injective. 
When high dimensional data, such as graphs, is involved, specific challenges arise, especially from the computational point of view. 

Kernel methods are considered to be among the most successful machine learning techniques for structured data. 
They replace the explicit projection in feature space with the evaluation of a symmetric semidefinite positive similarity function, called the kernel function. 
A major advantage of kernel methods is that very large, possibly infinite, feature spaces can be utilized by the learning algorithm with a computational burden dependent on the complexity of the kernel function and not on the size of the feature space. 
Unfortunately, any kernel function for graphs, whose correspondent feature space projection is injective, is as hard to compute as deciding whether two graphs are isomorphic \cite{Thomas2003}, which is believed to be a NP-Hard problem. 

As a consequence, in order to have computationally tractable kernel functions for graph data, a certain amount of information loss is inevitable. 
Most kernel functions for graphs associate specific types of substructures to features. The evaluation of the kernel function is then related to the number of common substructures between two input graphs. 
Such substructures include walks \cite{Kashima03marginalizedkernels} \cite{Vert2009} \cite{Vishwanathan2006}, paths \cite{Kriegel05shortestpath}  \cite{Heinonen2012}, specific types of subgraphs \cite{Costa2010} \cite{Shervashidze2009} and tree structures \cite{Martino2012}. Such kernels, with the exception of the ones in \cite{Martino2012} and \cite{Heinonen2012}, are computationally too demanding to be used with large datasets and  are effective when the correspondent features are relevant for the current task. 
Recently, the Fast Subtree Kernel has been proposed~\cite{NIPS2009_0533}. It has linear complexity (in the number of edges) and its features are subtree patterns of the input graphs. The kernel computes a rough approximation of the one-dimensional Weisfeiler-Lehman isomorphism test \cite{Weisfeiler1976}, with the explicit goal of being fast to compute. 

In this paper we present two kernel functions for graphs inspired by extensions of the Weisfeiler-Lehman isomorphism test. We define kernels whose feature space is much larger than the Fast Subtree Kernel with a modest increase in computational complexity.

\section{Weisfeiler-Lehman Isomorphism Test and Extensions} \label{sec:background}
Some notation is first introduced. 
A graph is a triplet $G = (V, E, L)$, where $V$ is the set of nodes and $|V|$ its cardinality, $E$ the set of edges and $L()$ a function returning the label of a node. 
A graph is undirected if $(v_i , v_j ) \in E \Leftrightarrow (v_j , v_i ) \in E$, otherwise it is directed. 
A path of length $n-1$ in a graph is a sequence of distinct nodes $v_1 ,\ldots, v_n$ such that $(v_i , v_i + 1)\in E$ for $1 \leq i < n$ ; if $v_1=v_n$ the path is a cycle. 
The distance $d(v_i,v_j)$ between the nodes $v_i,v_j$ is the length of any shortest path connecting them. 

We can now describe the  Weisfeiler-Lehman isomorphism test and a few extensions \cite{Cai1992} \cite{DeOliveiraOliveira2005}, which are all based on a relabelling process of the nodes of a graph $G=(V,E,L)$.  
We introduce two functions which, instantiated, determine the isomorphism test: $\pi(G, v)$, where $v\in V$, and $h()$ with the constraint that the codomain of $\pi(G, v)$ must coincide with the domain of $h()$. 
The role of $\pi(G, v)$ is to extract specific information from $G$: for example in the one-dimensional WL (1-dim WL) test $\pi(G, v)$ extracts the set of neighboring nodes of $v$: \mbox{$\pi(G, v)=\{u | u\in V, d(u,v)=1\}$}. 
The function $h()$ associates a unique numerical value (colour in the mathematical jargon) to each $\pi(G,v)$ and $h(\pi(G,v))$ will be used as novel label for $v$. 
In order for $h()$ to be well defined, a canonical representation for elements in its domain has to be defined, which practically boils down to defining a partial ordering between $\pi(G, v)$ elements. For example, in the 1-dim WL test the elements of $\pi(G, v)$ are sorted alphabetically according to their labels. 

The algorithm for computing the isomorphism test proceeds by iteratively relabelling $G$ nodes 
by means of a family of functions $L^{i}_{\pi}()$: 
\begin{equation} 
 L^{i}_{\pi}(v) = h(\pi(G^{i-1},v)), 
  \label{eq:L}
\end{equation}
where $G^0=(V,E,L)$ and $G^i=(V,E,L^{i}_{\pi})$ for $i>0$. 
The functions $L^{i}_{\pi}(v)$ are constructed for all $i\leq i^*$, where $i^*$ is the lowest index for which, $\forall v\in V$, $L^{i^*}_{\pi}(v) = L^{i^*+1}_{\pi}(v)$. Note that \mbox{$i^*\leq |V|$} for the 1-dim WL test~\cite{Miyazaki1996}. 
By applying the relabelling in eq.~\eqref{eq:L} to graphs $G$ and $G'$, we obtain two multisets of node labels: $\{L^{i^*}_{\pi}(v) | v\in V\}$ and $\{L^{j^*}_{\pi}(v') | v' \in V'\}$. If such multisets are different, then the two graphs 
are not isomorphic. On the contrary, if the two multisets are identical, there is not enough information to tell whether the two graphs are isomorphic. 

Extensions to the 1-dim WL test have been proposed to increase the discriminative power of the test. Their idea is to enrich the type of information used in the relabelling phase \cite{DeOliveiraOliveira2005}, \cite{Miyazaki1996}. The extension proposed by Miyazaki \cite{Miyazaki1996} considers the colour of the nodes up to distance $K$: $\pi(G, v)=\{(l, u) | u\in V, d(v,u)=l\leq K\}$; $\pi(G, v)$ elements, i.e. the tuples $(l, u)$, are ordered according to the relation $(l,u)<(l',u')\Leftrightarrow l<l' \vee $ $(l=l' \wedge L_\pi(u)<L_\pi(u'))$, where $L_\pi()$ is a generic labelling function. 
In the extension of Oliveira et al. \cite{DeOliveiraOliveira2005}, $h()$ is defined on paths, 
which are ordered according to the sequence of labels of the nodes in the path. 
Specifically, $\pi(G,v)$ extracts, for each $u\in V$, the shortest path between $v,u$ having lower $h()$ value: let $s(v,u)$ be the set of shortest paths connecting $u$ and $v$,  $\pi(G, v)=\cup_{u\in V} \arg\min_{p\in s(v,u)} h(p)$.

\section{Weisfeiler-Lehman kernel framework} \label{sec:framework}

Let us consider a function $\pi_r(G,v)$ depending on a parameter $r$, with $1\leq r \leq K$. 
Given a graph $G=(V,E,L)$, the application of eq.~\eqref{eq:L}, for a fixed $r$ value at the $i$-th iteration, yields the graph $G^i_r=G(V,E,L^i_{\pi_r})$, which differs from the original graph only in the labelling function. 
\begin{definition}\label{def:wlkernels}
Let $k()$ be any kernel for graphs that we will refer to as the base kernel. 
Then the {\it Extended Weisfeiler-Lehman kernel} with $h$ iterations, depth $K$ and base kernel $k()$ is defined as:
\begin{equation}
WL^{K}_h (G,G')= \sum_{r=1}^K \sum_{i=0}^h k(G^i_r,G^{'i}_{r}).
\label{eq:kdimwlkernel}
\end{equation}
\end{definition}
Since the functions in eq.~\eqref{eq:L} are well defined and the Extended Weisfeiler-Lehman kernel of eq.~\eqref{eq:kdimwlkernel} is a finite sum of positive semidefinite functions, it is also positive semidefinite. 

\begin{figure}[t]
\centering
\begin{tikzpicture}[auto,thick,scale=.5]
   \tikzstyle{graph}=[scale=1.0,minimum size=12pt, inner sep=0pt, outer sep=0pt, draw, circle, text=black]
		\node (s) [graph] {s}   
					child {
            node (b) [graph] {b}
        	};
         \node (e) [graph] at (0:2){e}   
         		child { node (d) [graph] {d}};		
    ;    	
    \draw
    (s) to (e)
    (b) to (d)
    (b) to (e)
    ;
  \small
   \tikzstyle{node}=[draw, minimum size=12pt, inner sep=0pt, outer sep=0pt, circle,text=black]
        \node [node,right=2.25cm of s, yshift=0.3cm] (dag) {\textbf{s}} [->] 
                child {
                    node [node] {\textbf{e}}[] 
                    child {node (d2)[node, xshift=0.35cm]{\textbf{d}}[]}
                }
                child {
                    node (b2) [node]{\textbf{b}}[]
                }
				;
         \draw [->]
         (b2) to (d2)
        ;
        \node (treevisit) [node,right=1.5cm of dag, yshift=0cm] {\textbf{s}} [->] 
                child {
                    node [node] {\textbf{e}}[] 
                    child {node [node]{\textbf{d}}[]}
                }
                child {
		    node (b2) [node]{\textbf{b}}[]	
		    child {node [node]{\textbf{d}}[]}	
                }
	;
        \node (feature1) [node,right=1.4cm of treevisit, yshift=-1.5cm] {\textbf{d}};
        \node (feature2) [node,right=2.4cm of treevisit, yshift=-1.5cm] {\textbf{d}};
        \node (feature3) [node,right=1.4cm of treevisit, yshift=0cm] {\textbf{e}} [->] child {node [node] {\textbf{d}} []};
        \node (feature4) [node,right=2.4cm of treevisit, yshift=0cm] {\textbf{b}} [->] child {node [node] {\textbf{d}} []};
        \node (feature5) [node,right=3.7cm of treevisit, yshift=0cm] {\textbf{s}} [->] 
                child {
                    node [node] {\textbf{e}}[] 
                    child {node [node]{\textbf{d}}[]}
                }
                child {
		    node (b2) [node]{\textbf{b}}[]	
		    child {node [node]{\textbf{d}}[]}	
                }
	;
    \tikzstyle{node}=[text=black]
    \node (la) [node, left=0.15cm of s, yshift=-0.45cm] {a)} [];
    \node (lb) [node, right=1.85cm of la] {b)} [];
    \node (lc) [node, right=1.45cm of lb] {c)} [];
    \node (ld) [node, right=1.65cm of lc] {d)} [];
\end{tikzpicture}
\caption{Steps for obtaining some of the features of the $WL_{DDK}$ kernel: a) an input graph $G$; b) the DAG resulting from the application of $\pi(G, v)$ where $v$ is the node labelled as \textbf{s}; c) the tree visit $T(v)$; d) the features of the ST kernel related to $T(v)$.\label{fig:wlddk}}
\end{figure}
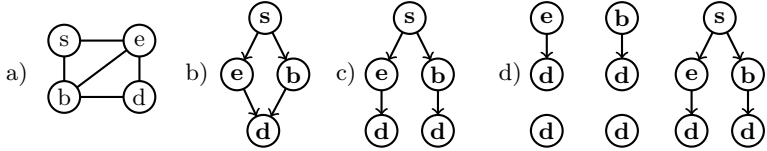

Let us now present the main contribution of the paper, i.e. two novel kernels which are instances of eq.~\eqref{eq:kdimwlkernel}. 
For both kernels the function $\pi(G, v)$ returns the following Directed Acyclic Graph (DAG) rooted at $v$: $D_r(v)=(V_r,E_r,L)$ where $V_r=\{u\in V | d(v,u)\leq r\}$ and $E_r$ consists in all edges of $G$ that appear in any of the shortest path connecting $v$ and any $u\in V_r$ (see  Fig.~\ref{fig:wlddk}-b for an example). 
In order to have a canonical representation for the DAG $D_r(v)$, the ordering for DAG nodes described in \cite{Martino2012} is used.
The function $h()$ assigns a unique numerical value to each DAG, and it can be implemented efficiently as presented in~\cite{DaSanMartino2012}.
Let the maximum number of nodes of each DAG $D_r(v)$ be $|D_r|$. 
 Then it can be shown that $|D_r|$ is $O(\rho^{r})$ \cite{Martino2012}, where $\rho$ is the maximum node outdegree. 
Computing all the indices $L^i_{\pi_r}()$ for a graph $G$ has worst-case time complexity $O(|D_r||V|\log |D_r||V|)$ (see  \cite{Martino2012} for details). Assuming $\rho$ constant (a condition that usually holds in real-world datasets) the worst-case time complexity reduces to $O(|V|\log |V|)$.

%

In the first proposed kernel, that we will refer to as $WL_{NS-DDK}$, the base kernel is defined as 
\begin{equation}
 k(G_{r}^i,G_{r}^{'i})=\sum_{v \in V} \sum_{v' \in V'}\delta(L^i_{\pi_r}(v),L^i_{\pi_r}(v')), 
  \label{eq:k1}
\end{equation}
 where $\delta$ is the Kronecker's delta function. Note that computing the kernel is equivalent to performing a hard match between the DAGs encoded by $L^i_{\pi_r}(v)$ and $L^i_{\pi_r}(v')$. If we order the list of indices $\{L^i_{\pi_r}(v) | v\in V\}$ and $\{L^i_{\pi_r}(v') | v'\in V'\}$, then eq.~\eqref{eq:k1} can be computed in 
 $O(|V|\log |V|)$ time.

The second kernel we propose, referred to as $WL_{DDK}$, differs from the first one only in the base kernel $k()$. 
Let $T(v)$ be the function that, first computes the DAG $\pi_r(G,v)$ and then returns the tree
resulting from the breadth-first visit of the DAG starting from $v$ (see Fig.~\ref{fig:wlddk}-c for an example). 
Finally, $k()$ can be defined as any kernel for trees applied to $T(v)$ and $T(v')$, for example the subtree kernel (ST) \cite{DBLP:conf/nips/ViswanathanS02}: 
\begin{equation}
 k(v,v') = \sum_{v\in V}\sum_{v'\in V'} k_{ST}(T(v),T(v')). 
  \label{eq:k2}
\end{equation}
The ST kernel counts the number of matching proper subtrees of $T(v)$ and $T(v')$, where a proper subtree of a tree $T$ rooted at $u$ is the subtree composed by $u$ and all of its descendants (in Fig.~\ref{fig:wlddk}-d are listed the set of proper subtrees of the tree in Fig.~\ref{fig:wlddk}-c). The complexity of $k_{ST}(T,T')$ is $O(n\log n)$ where $n=\min(|T|,|T'|)$. 
Assuming $\rho$ constant, $O(|T(v)|)=O(|D_r(v)|)$. By using the algorithm described in \cite{Martino2012}, the complexity of computing eq.~\eqref{eq:k2} is $O(|V|\log |V|)$. 

There are a number of kernels in literature that are instances of eq.~\eqref{eq:kdimwlkernel}. 
The Fast Subtree Kernel (FS) counts the number of identical subtree patterns of depth $h$ \cite{NIPS2009_0533}. 
It can be obtained from eq.~\eqref{eq:kdimwlkernel} by setting: {\it i)} $K=1$; \mbox{ {\it ii)} $\pi(G,v)=\{u|u\in V, d(v,u)=1\}$} and then ordering $\pi(G,v)$ elements alphabetically according to their labels; {\it iii)} the base kernel $k()$ is the one in eq.~\eqref{eq:k1}. 
The ODD-ST$_h$, described in \cite{Martino2012}, is an instance of the $WL_{DDK}$ of eq.\eqref{eq:k2} and it is obtained setting $h=0$ in  eq.~\eqref{eq:kdimwlkernel}. 

\section{Experimental results} \label{sec:exps}

In this section, we compare the two kernels presented in Section~\ref{sec:framework} against other state-of-the-art kernels on five real-world datasets.

 We considered the Fast Subtree kernel~\cite{NIPS2009_0533}, the ODD-ST$_h$ kernel~\cite{Martino2012} (described in section~\ref{sec:framework}) and the NSPDK kernel~\cite{Costa2010} , that computes the exact matches between pairs of subgraphs with controlled size and distance. 
 For the assessment of the performance of the proposed kernels, we considered five real-world datasets: CAS\footnote{http://www.cheminformatics.org/datasets/bursi}, CPDB~\cite{Helma2004}, AIDS~\cite{Weislow1989}, NCI1~\cite{springerlink:10.1007/s10115-007-0103-5} and GDD~\cite{dobson2003}. 
All the datasets represent binary classification problems. The first four datasets involve chemical compounds, represented as graphs where the nodes represent the atoms (labelled according to the atom type) and the edges the bonds between them. In chemical compounds, there are no self-loops. 
GDD is a dataset of proteins,  where each protein is represented by a graph, in which the nodes are amino acids and two nodes are connected by an edge if they are less than $6^\circ$  Angstroms apart. 
CAS and NCI1 are the largest datasets, with $4337$ and $4110$ examples, respectively. For more information about the datasets, please refer to~\cite{Martino2012}. 
%

All the kernels have been employed together with a Support Vector Machine. The $C$ parameter of the SVM has been selected in the set $\{0.01,0.1,1,10,100\}$.
For all the experiments, the values of the parameters of the ODD-ST$_{h}$ kernel have been restricted to: $h=\{1, 2, \ldots, 8\}$ \mbox{$\lambda=\{0.1, 0.2, \ldots, 2.0\}$} ($\lambda$ is a parameter of $K_{ST}$);
for the Fast Subtree Kernel we optimized  the only parameter of the kernel $h=\{1, 2, \ldots, 10\}$ ; for the NSPDK kernel we optimized the parameters  $r=\{1, 2, \ldots, 8\}$ and $d=\{1, 2, \ldots, 8\}$.
Concerning the two kernels presented in this article, their parameters are $K=\{1, 2, 3, 4\}$ , $h=\{0,1, 2, \ldots, 8\}$ and $\lambda=\{0.1, 0.2, \ldots, 2.0\}$.
The parameters range has been selected in such a way that the computational time needed for the calculation of the kernel matrices is roughly comparable, i.e. at most one hour on a modern PC. 
For parameter selection we adopt a technique commonly referred to as \textit{nested} K-fold cross validation following ~\cite{NIPS2009_0533}. 
All the experiments have been repeated $10$ times and the average results (with standard deviation) are reported.
 \begin{table}[t]
\centering
  \begin{tabular}{|l|c|c|c|c|c|c|}
\hline
$Kernel$ 		  	& CAS 		& CPDB 		& AIDS 		& NCI1  	& GDD & AVG Rank \\
\hline
FS			 	& $81.05$ (5) 	& $73.22$ (5)	& $75.61$ (5)	& $84.77$ (3)	& $76.21$ (2) & 4 \\
			& \tiny{\raisebox{0.2cm}{($\pm 0.50$)}}& \tiny{\raisebox{0.2cm}{($\pm 0.78$)}}& \tiny{\raisebox{0.2cm}{($\pm 1.00$)}}&  \tiny{\raisebox{0.2cm}{($\pm 0.31$)}} & \tiny{\raisebox{0.2cm}{($\pm 1.15$)}}& \\
		
NSPDK			 	& $83.60$ (2)	& $76.99$	(2)  	& $82.71$ (3) 	& $83.46$ (4)	& $74.09$ (5) & 3.2\\

			& \tiny{\raisebox{0.2cm}{($\pm 0.34$)}}& \tiny{\raisebox{0.2cm}{($\pm 1.15$)}}& \tiny{\raisebox{0.2cm}{($\pm 0.66$)}}&  \tiny{\raisebox{0.2cm}{($\pm 0.46$)}} & \tiny{\raisebox{0.2cm}{($\pm 0.91$)}}& \\
$ODD-{ST_h}$	 	&  $83.34$(3)	& $76.44$	 (4)  	& $81.51$(4) 	& $82.10$ (5)	& $75.23$(4) & 4 \\

			& \tiny{\raisebox{0.2cm}{($\pm 0.31$)}}& \tiny{\raisebox{0.2cm}{($\pm 0.62$)}}& \tiny{\raisebox{0.2cm}{($\pm 0.74$)}}&  \tiny{\raisebox{0.2cm}{($\pm 0.42$)}} & \tiny{\raisebox{0.2cm}{($\pm 0.70$)}}& \\

$WL_{NS-DDK}$		 	& $82.96$ (4)	& \underline{$77.03$}	 (1)  	& $82.80$ (2) 	& $84.79$ (2)	& \underline{$77.20$} (1)& 2 \\

			& \tiny{\raisebox{0.2cm}{($\pm 0.49$)}}& \tiny{\raisebox{0.2cm}{($\pm 1.18$)}}& \tiny{\raisebox{0.2cm}{($\pm 0.66$)}}&  \tiny{\raisebox{0.2cm}{($\pm 0.36$)}} & \tiny{\raisebox{0.2cm}{($\pm 0.65$)}} & \\
			$WL_{DDK}$	 	& \underline{$83.91$} (1)	& $76.52$	 (3)  	& \underline{$82.93$}(1) 	& \underline{$84.90$} (1)	&$75.45$ (3) & 1.8 \\

			& \tiny{\raisebox{0.2cm}{($\pm 0.29$)}}& \tiny{\raisebox{0.2cm}{($\pm 1.16$)}}& \tiny{\raisebox{0.2cm}{($\pm 0.71$)}}&  \tiny{\raisebox{0.2cm}{($\pm 0.33$)}} & \tiny{\raisebox{0.2cm}{($\pm 0.86$)}} & \\

\hline
\end{tabular}
\vspace{0.2cm}
\normalsize
 \caption{Average accuracy results $\pm$ standard deviation in nested 10-fold cross validation for the Fast Subtree, the Neighborhood Subgraph Pairwise Distance, the $K_{ODD-ST_h}$, $WL_{NS-DDK}$ and $WL_{DDK}$ kernels obtained on CAS, CPDB, AIDS, NCI1 and GDD datasets. The rank of the kernel is reported between brackets. \label{tab:nestedkfoldresults}}
\end{table}
\begin{figure}[t]
\centering
  \includegraphics[width=0.8\textwidth]{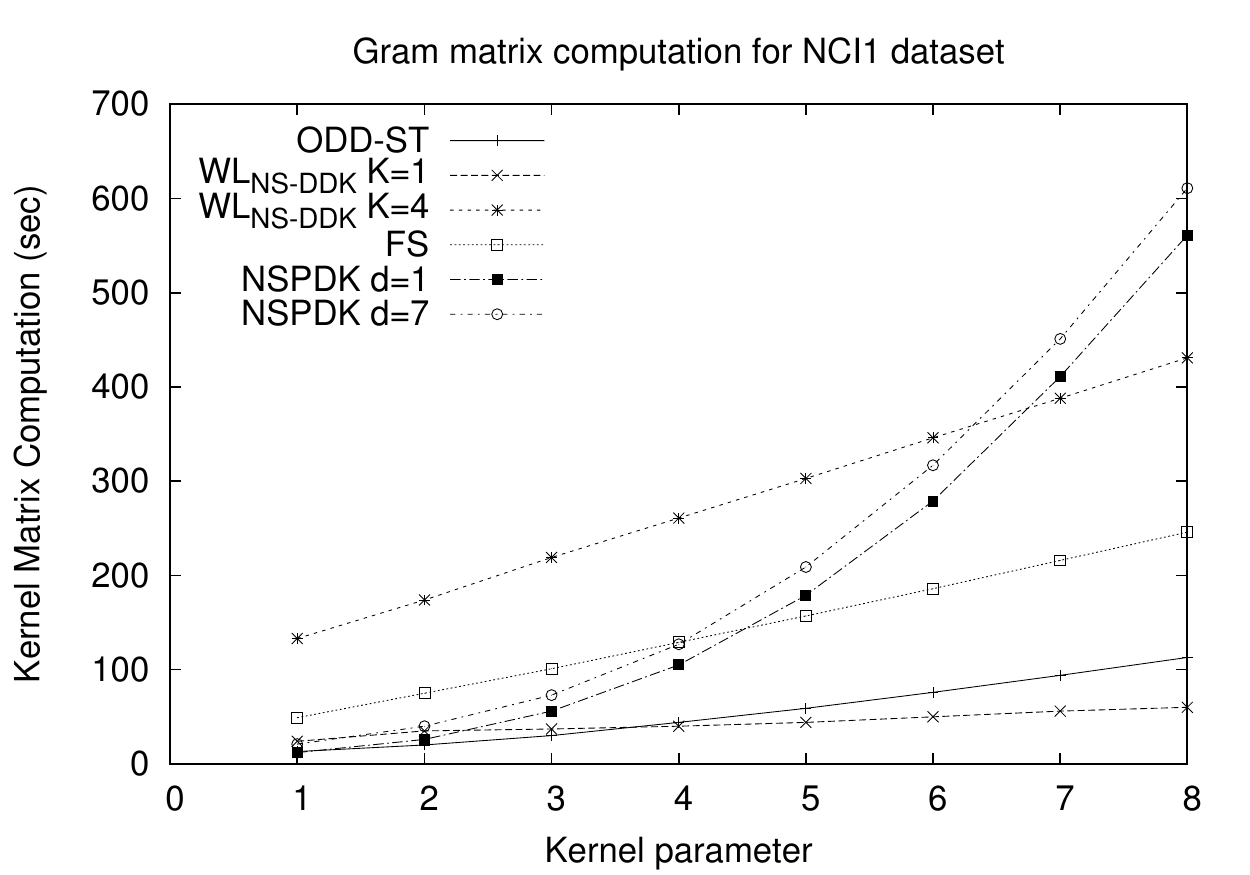}
  \caption{Comparison between the time needed for computing the Gram matrix on the NCI1 dataset for the different kernels, as a function of the parameter: $h$ for FS and $WL_{NS-DDK}$, $K$ for $ODD-ST$, $r$ for NSPDK.\label{fig:timesNCI}}
\end{figure}

Table~\ref{tab:nestedkfoldresults} summarizes the average accuracy results of the proposed kernels and the state-of-the-art ones on the considered datasets. The mean accuracy is reported with the standard deviation. Between brackets, the ranking of the specific kernel on the dataset is reported. In the rightmost column, the average ranking value on all the datasets for each kernel is reported. 
When considering single datasets, there is no dataset where NSPDK or FS kernels rank first.
On all the considered datasets, either $WL_{NS-DDK}$ or $WL_{DDK}$ outperforms the other kernels. 
If we look at the average ranking, the situation is clearer. 
The best average ranking of the competing kernels is the one of $NSPDK$, with a value of $3.2$.
The $WL_{NS-DDK}$ has an average ranking of $2$. $WL_{DDK}$ performs slightly better, with an average ranking value of $1.8$.
These results clearly show that, on the considered datasets, the $WL$ kernel family performs better than the other kernels present in literature. 

Figure~\ref{fig:timesNCI} reports the computational time, in seconds, needed from the \\$WL_{NS-DDK}$ kernel and the competing ones to compute the Gram matrix for the NCI1 dataset. The computation time required by $WL_{DDK}$ is very similar and thus omitted.
\section{Conclusions and future work} \label{sec:conclusions}
This paper proposed a new framework for the definition of graph kernels based on a generalization of the 1-dimensional WL test.
The framework can be instantiated with any kernel for graphs as a base kernel. In particular, we analyzed two instances inspired by the Decompositional DAGs graph kernels \cite{Martino2012}.
The two kernels show state-of-the-art predictive performance on five real-world datasets, with a computational burden that, on such datasets, grows only linearly with respect to the kernel parameters. 
As a future work, we will explore other members of the framework.
\subsubsection{Acknowledgments.} This work was supported by the University of Padova under the strategic project BIOINFOGEN.

\bibliographystyle{abbrv}
\bibliography{Mendeley}
\end{document}